\documentclass{article}

\usepackage[preprint]{neurips_2025}

\usepackage[utf8]{inputenc} % allow utf-8 input
\usepackage[T1]{fontenc}    % use 8-bit T1 fonts
\usepackage[colorlinks]{hyperref}       % hyperlinks
\usepackage{url}            % simple URL typesetting
\usepackage{booktabs}       % professional-quality tables
\usepackage{amsfonts}       % blackboard math symbols
\usepackage{nicefrac}       % compact symbols for 1/2, etc.
\usepackage{microtype}      % microtypography
\usepackage{xcolor}         % colors
\usepackage{tikz}
\usepackage{color}
\usepackage{pgfplots}
\pgfplotsset{compat=1.18}
\usetikzlibrary{positioning}
\usepackage{xspace}

\title{4Hammer: a board-game reinforcement learning environment for the hour long time frame}

\author{%
  Massimo Fioravanti\\
  DEIB -- Politecnico di Milano\\
  Via G. Ponzio 34/5, I-20133 Milano, Italy\\
  \texttt{massimo.fioravanti@polimi.it} \\
  \And
  Giovanni Agosta\\
  DEIB -- Politecnico di Milano\\
  Via G. Ponzio 34/5, I-20133 Milano, Italy\\
  \texttt{giovanni.agosta@polimi.it} \\
}

\newcommand{\Rulebook}{\emph{Rulebook}\xspace}

\newcommand{\CPP}{\texttt{C++}\xspace}

\newcommand{\Hammer}{\emph{4Hammer}\xspace}
\newcommand{\RealHammer}{\emph{Warhammer 40.000}\xspace}

\begin{document}

\maketitle

\begin{abstract}
Large Language Models (LLMs) have demonstrated strong performance on tasks with short time frames, but struggle with tasks requiring longer durations. While datasets covering extended-duration tasks, such as software engineering tasks or video games, do exist, there are currently few implementations of complex board games specifically designed for reinforcement learning and LLM evaluation. To address this gap, we propose the \Hammer reinforcement learning environment, a digital twin simulation of a subset of \RealHammer—a complex, zero-sum board game. \RealHammer features intricate rules, requiring human players to thoroughly read and understand over 50 pages of detailed natural language rules, grasp the interactions between their game pieces and those of their opponents, and independently track and communicate the evolving game state.
\end{abstract}

\section{Introduction}
%\todo{talk about llms}

Machine Learning (ML) techniques have been able to tackle tasks with progressively longer time horizons \cite{kwa2025measuringaiabilitycomplete}, yet many tasks commonly performed by humans are still beyond the abilities of machines to perform. Games have long attracted research interest in artificial intelligence \cite{CAMPBELL200257, silver2016mastering, shao2019survey} as they provide structured environments with clearly defined rules within which an agent can perform actions as well as variable time horizon length. The two main categories of games relevant to ML are video games~\cite{bergdahl2020augmenting,shao2019survey,souchleris2023reinforcement} and board games~\cite{szita2012reinforcement,xenou2018deep,perolat2022mastering,patankar2024survey}. The former relieve players of the mental burden of tracking the game state by providing user interfaces that present necessary information on-demand and enforce valid player actions. In contrast, board games typically provide physical components for maintaining game states, yet frequently leave tracking transient elements—such as turn order—to the players themselves, thus increasing the difficulty of playing them for an ML agent.

It is worth noting that board games are not an homogeneous class. Abstract games, such as chess and go, have simple rules but emerging complexity, while other games have an increased cognitive cost arising from large amount of rules and components -- f.i., Magic: The Gathering has hundreds of different cards, each presenting its own tiny set of rules which are added to the baseline rules of the game, while \RealHammer has many different pieces (miniatures) which behave according to their own rules variations. For human players, physical game components (e.g., cards reporting the rules) and/or on intangible qualities, such as their theme, are used to help in dealing with this complexity \cite{garfield}. While abstract games offer a precise rules environment, thanks to their abstraction from reality, some commercial board games exhibit rules that are informal, incomplete, contain inaccuracies or implicitly rely on the player's understanding of the real world.

Consequently, learning and playing a new off-the-shelf board game from start to finish constitutes a complex, interactive human task that can span multiple hours, requiring players to navigate environments characterized by rule uncertainty and sometime even resolve disputes arising from differing interpretations of the game state.

It is worth noting that large amounts of rules make digital implementations of such games difficult, therefore large non-abstract board games digital implementations targeting ML are few. Many ML projects only offer abstract board games that are easy to implement \cite{LanctotEtAl2019OpenSpiel,Piette_2020_Ludii,patankar2024survey}. 

In this paper we introduce \Hammer, the first digital implementation of \RealHammer \cite{gw-40k}, a off-the-shelf board with large amounts of textual and situational rules and player tracked game state. Although \RealHammer is not as universally known as chess or poker, \RealHammer is by far the most widely played game in the category of miniature war games, with at least 300.000 games played in two-day tournaments since August 2023 \cite{statcheck}. Thus, an implementation amenable to ML would constitute an important stepping stone towards the automated solution of games with large rules sets. 

\Hammer can be used both with and without a graphical engine to render the top down view of the game state, can serialize textual and tensor representations of the game state and thus be used both with LLMs and traditional reinforcement learning techniques. We demonstrate both capabilities with an experimental campaign. 

The rest of this paper is organized as follows. 
In section~\ref{sec:bg}, we briefly introduce \RealHammer and its ``Combat Patrol'' variant, which is the focus of the proposed implementation.
In section~\ref{sec:arch}, we explain the architecture of the \Hammer implementation and its integration with graphical engines, LLMs, and Reinforcement Learning libraries, while in section~\ref{sec:eval} we provide an experimental assessment.
Finally, in section~\ref{sec:conc} we draw some conclusions and highlight future directions.

\section{Background}
\label{sec:bg}

\paragraph{Warhammer 40,000} \RealHammer  is a non-deterministic, two-player, zero-sum tactical miniature board game. Each player selects various game pieces, known as \emph{units}, each possessing distinct rules and abilities. These units are deployed onto a three-dimensional battlefield, where players alternate turns to maneuver their units strategically, engage opposing forces, and complete specific objectives.

\RealHammer presents several characteristics that pose challenges for machine learning approaches:

\begin{itemize} \item \textbf{Extended Timeframe:} A single match can span several hours, with tournament rounds typically lasting between one and three hours. While part of this time is devoted to physically moving game pieces on the board, which does not happen in a digital implementation, the time spent in decision-making is still much longer than in abstract games.

\item \textbf{Extensive Core Rules:} The core rules are detailed in a freely available, 60-page natural-language document, supplemented by an additional 35-page document covering advanced game mechanics. Not all rules apply in every match; f.i., rules related to vehicles might be irrelevant in games where no vehicle-type units are present.

\item \textbf{Extensive Faction-Specific Rules:} The complete game offers hundreds of unit options, each possessing unique rules. Human players are typically expected to know the rules governing their chosen units and must communicate these rules to opponents upon request.

\item \textbf{Player-Tracked Game State:} Many game effects persist across multiple actions or turns. While some rules explicitly instruct players on how to track such effects—such as placing specific tokens—other temporary effects require players to independently monitor and remember the current state. E.g., units typically move only once per turn, and it is the player's responsibility to remember which units have already moved.
\end{itemize}

\paragraph{Combat Patrol:} \RealHammer includes several game variants, called \emph{modes}, one of which, Combat Patrol, serves as an introductory variant. This mode significantly reduces the complexity by limiting the combinations and total number of units that players select before gameplay, although the core rules remain unchanged. 

\section{The \Hammer environment}

The \Hammer environment is a digital implementation of \RealHammer's Combat Patrol game mode. Its rules are encoded in \Rulebook\footnote{\href{anonymous repo}{https://github.com/rl-language/rlc/}}, a domain-specific language explicitly designed for creating reinforcement learning datasets. The graphical interface of the environment is powered by the Godot game engine \cite{godot-engine}, released under MIT license.

The \Hammer environment is designed recognizing that the field of ML is in continuous flux, and thus environments should support multiple use cases. In particular, we identified the following use cases, which can be combined as needed:

\begin{itemize} 
\item \textbf{Perfect Information Reinforcement Learning:} The entire game state is directly accessible as observations provided to the reinforcement learning algorithm.

\item \textbf{Imperfect Information Reinforcement Learning:} The game state is observed indirectly via a rendered, bird's-eye graphical view of the game board. Agents must independently track and maintain additional information that isn't explicitly provided.

\item \textbf{State with Rules:} The full game rules are accessible, and agents select actions from a provided list of valid options.

\item \textbf{State Only:} The formal rules implementation is disregarded, and agents directly manipulate the underlying state data structure as a working memory (``scratch pad'') to carry out actions in the game.

\end{itemize}

\textbf{Implemented systems}
The \Hammer environment currently implements 6 of the 35 available Combat Patrol factions, comprising a total of 22 distinct units, along with all necessary rules to use these units. This simplification does not detract significantly from the game, since the factions are never used all at the same time -- rather, each player chooses a different faction.

\subsection{Limitations}
Rules not directly relevant to these selected units have not been implemented. Additionally, the physical game is originally played on a 44x30-inch board, where units can occupy continuous 3D positions. In \Hammer, we have discretized the board into 1-inch squares, modifying movement rules accordingly to accommodate this quantization. The original three-dimensional environment has also been simplified into a two-dimensional representation, removing obstacles and rendering all game elements as two-dimensional objects.
Moreover, the optional secondary objective mechanics have been omitted from our implementation. 

The omitted rules interfere with the game play at the tactical level where minute distances of where models are placed on the table are relevant but have little effect on the strategic level of the game, which focuses more on selecting the right game components to achieve objectives such as trying to eliminate the opponent game pieces. In our experiments, we have not observed neural networks or LLMs achieving good results at the strategic level; therefore, the finer tactical level is irrelevant. When the models will be able to perform well on the strategic level, the tactical level can be expanded to include all rules.

\section{Architecture}
\label{sec:arch}

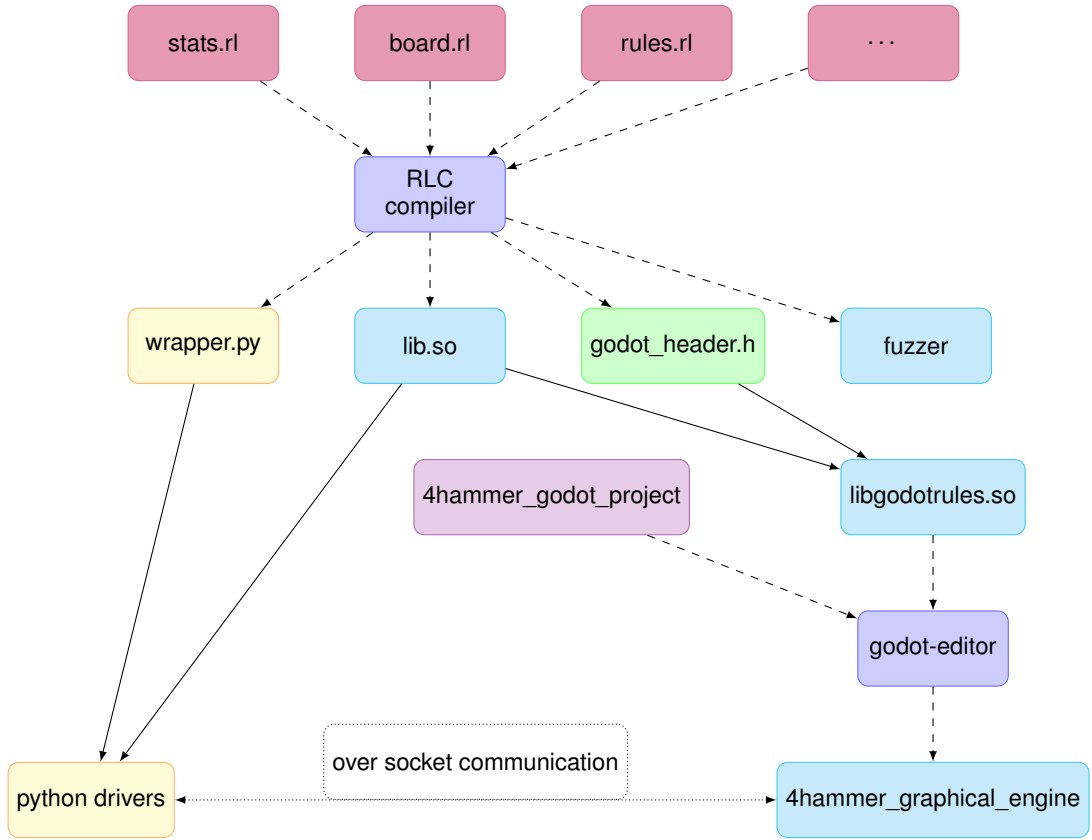
\begin{figure}
    \centering

\begin{tikzpicture}[
  every node/.style   = {rectangle, draw, rounded corners,
                         minimum width=2cm, minimum height=1cm,
                         align=center, font=\sffamily\small},
  purplebox/.style    = {fill=purple!40,  draw=purple!60},
  bluebox/.style      = {fill=blue!20,    draw=blue!60},
  redbox/.style       = {fill=cyan!20,     draw=cyan!60},
  orangebox/.style    = {fill=yellow!20,  draw=orange!60},
  greenbox/.style     = {fill=green!20,   draw=green!60},
  violetbox/.style    = {fill=violet!20,  draw=violet!60},
  dashededge/.style   = {dashed, -latex},
  solidedge/.style    = {-latex},
  socketedge/.style   = {densely dotted}   % <-- no arrow here; we add it when drawing
]

% ───────────────── Row 1: purple *.rl inputs ─────────────────
\node[purplebox]                          (stats)  {stats.rl};
\node[purplebox, right=1cm of stats]      (boar)   {board.rl};
\node[purplebox, right=1cm of boar]       (rules)  {rules.rl};
\node[purplebox, right=1cm of rules]       (otherrl)  {\dots};

% ───────────────── Compiler ─────────────────
\node[bluebox, below=1cm of boar]         (rlc)    {RLC\\\footnotesize compiler};

\foreach \src in {stats,boar,rules,otherrl}{\draw[dashededge] (\src) -- (rlc);}

% ───────────────── Compiler outputs (wrapper ↔ lib.so swapped) ─────────────────
\node[orangebox, below left =1cm and 1cm of rlc] (wrapper) {wrapper.py};
\node[redbox,    below      =1cm         of rlc]   (libso)   {lib.so};
\node[greenbox,  below right=1cm and 1cm of rlc] (header)  {godot\_header.h};
\node[redbox,    right=1cm  of header]               (fuzzer)  {fuzzer};

\foreach \dst in {libso,wrapper,header,fuzzer}{\draw[dashededge] (rlc) -- (\dst);}

% ───────────────── libgodotrules.so ─────────────────
\node[redbox, below right=1cm and 1cm of header] (godotrules) {libgodotrules.so};
\draw[solidedge] (header) -- (godotrules);
\draw[solidedge] (libso)  -- (godotrules);

% ───────────────── 4hammer_godot_project (lifted up) ─────────────────
\node[violetbox, left=2cm of godotrules] (project) {4hammer\_godot\_project};

% ───────────────── Godot editor & engine ─────────────────
\node[bluebox, below=1cm of godotrules]                   (editor)  {godot-editor};
\node[redbox, below=1cm of editor]                       (engine)  {4hammer\_graphical\_engine};

\draw[dashededge] (project)    -- (editor);
\draw[dashededge] (godotrules) -- (editor);
\draw[dashededge]  (editor)     -- (engine);

% ───────────────── python drivers (lowered to engine level) ─────────────────
\node[orangebox, left=8cm of engine] (drivers) {python drivers};

\draw[solidedge] (libso)   -- (drivers);   % vertical+horizontal to new position
\draw[solidedge] (wrapper) -- (drivers);

% ───────────────── Bidirectional socket connection ─────────────────
\draw[socketedge, latex-latex] (engine) -- node[midway, above] {over socket communication} (drivers);

\end{tikzpicture}

   \caption{Hierarchical representation of the components in \Hammer. Purple components are implemented exclusively in \Rulebook, while yellow components represent Python source files. The component labeled \texttt{4hammer\_godot\_project} comprises Godot engine files responsible for graphical elements. Cyan-colored nodes represent binary objects such as executables and libraries. Dashed edges indicate input-output relationships, illustrating inputs processed by tools (\texttt{rlc} or the Godot editor) to produce outputs. Solid lines represent composition, illustrating how various artifacts are combined to produce new components. Dotted lines indicate network communication occurring at runtime.}
    \label{fig:architecture}
\end{figure}

Figure \ref{fig:architecture} shows the architecture of \Hammer from a build system point of view. The game simulation libraries are compiled into a shared library, and wrappers for Python, \CPP and Godot are generated. From the rules library, the handwritten graphical godot components we build the graphical engine. Python script can instead dynamically load the rules library. \Hammer can be found at https://github.com/rl-language/4Hammer . 

\subsection{Game simulation libraries}

The Stats, State, and Rules are the core simulation components and contain the code to run it without the graphical engine. They supports bidirectional interoperability with Python, C and C++, provide serialization to tensor encoding, textual encoding and binary encoding, and native handling of hidden information. These components are implemented in \Rulebook, a domain-specific language that provides the mentioned features.

Each of these libraries is modular, depending only on required components, and can therefore be independently reused.

\paragraph{Stats} The Stats library contains declarative data for all units implemented within \Hammer. This library can easily be expanded by adding rules for additional units.

\paragraph{Board} The Board library defines all essential elements needed to represent the game state, excluding game sequences. All scalar values and containers within this library are annotated with their minimum and maximum possible values, enabling validation of game states. Being written purely in \Rulebook, textual, binary, and tensor representations are automatically generated. Consequently, the game state can be seamlessly provided either to LLMs through a textual representation or used directly with traditional reinforcement learning methods.

\begin{figure}[htbp]
  \centering
  \begin{minipage}{0.48\linewidth}
    \includegraphics[width=\linewidth]{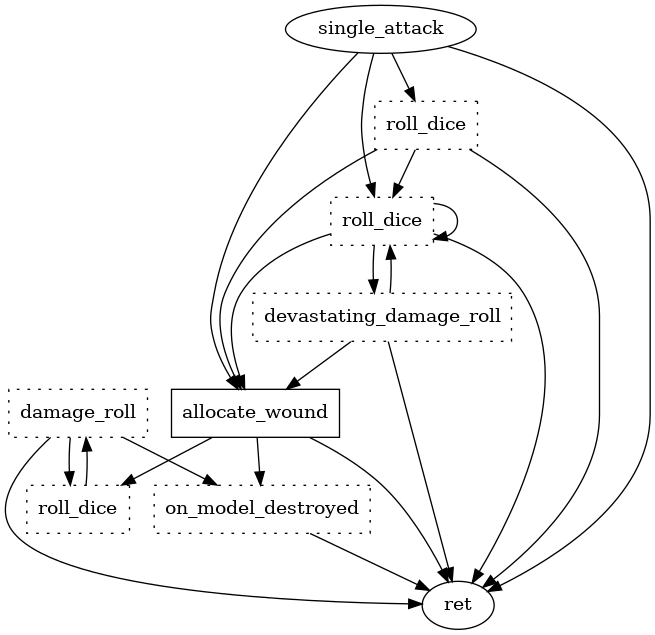}
  \end{minipage}
  \hfill
  \begin{minipage}{0.48\linewidth}
    \includegraphics[width=\linewidth]{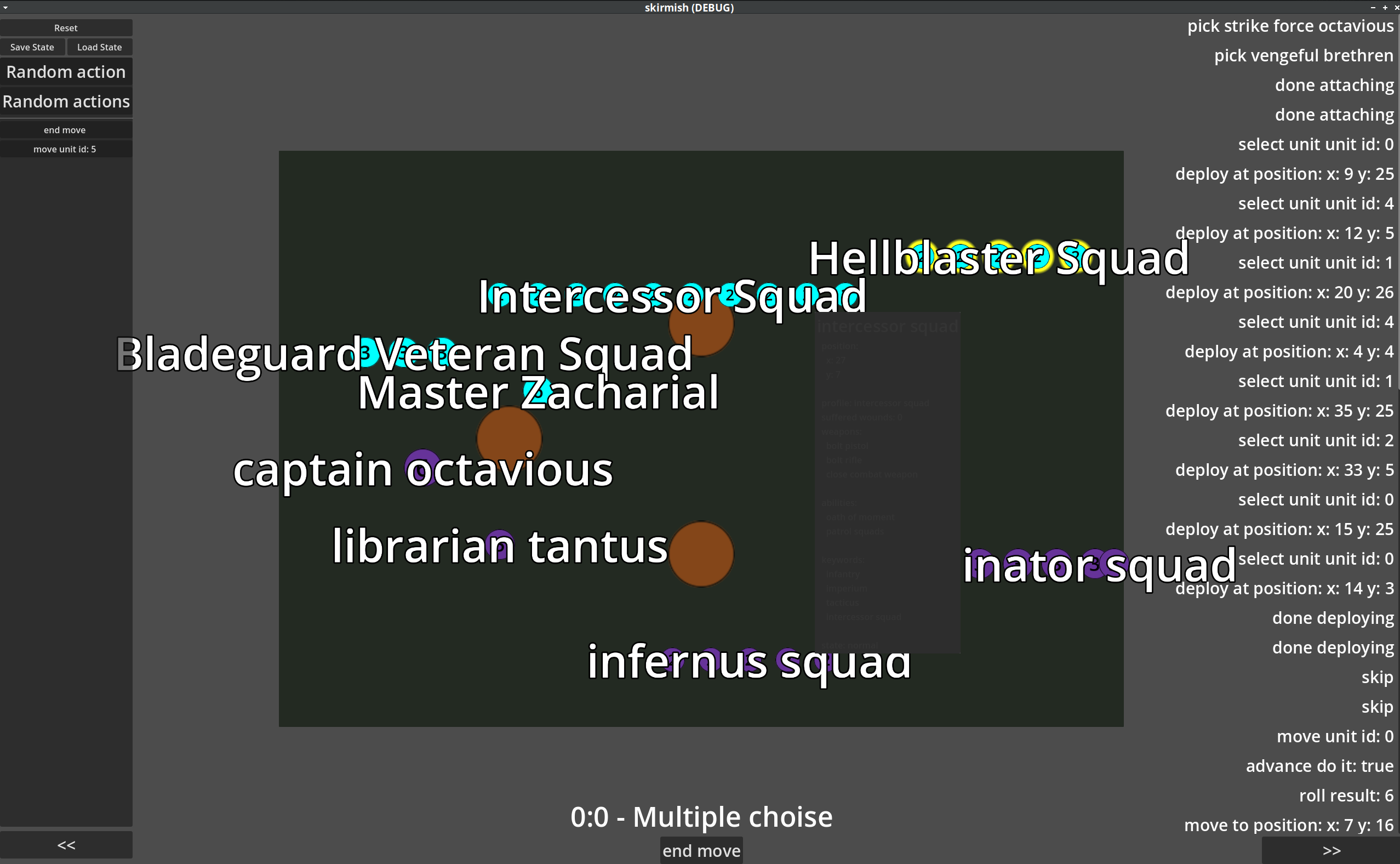}
  \end{minipage}

  \caption{The single\_attack game sequence (left) and the engine‐level skirmish (right).}
  \label{fig:two-side-by-side}
\end{figure}

\paragraph{Rules} The Rules library encapsulates all gameplay sequences required to play a complete match from start to finish. For example, Figure \ref{fig:two-side-by-side} illustrates the control flow graph for the single\_attack sequence, executed potentially hundreds of times per game when a player's unit interacts with opposing units. Ellipses represent entry and exit points of sequences, solid boxes indicate atomic decisions made by players (e.g., allocating damage to game components), and dashed boxes represent nested sequences called within the current sequence. Arrows indicate potential transitions between nodes based on the game's state during simulation.

Sequences are reusable and composable, as demonstrated by Figure \ref{fig:two-side-by-side}, where the roll\_dice subsequence is utilized multiple times. This reusability facilitates the composition of new environments beyond those described in this document, for example, different game modes.

\paragraph{Compilation Outputs} From the \Rulebook source files, several outputs are automatically generated: a library implementing the rules, a fuzzer for automated testing, a Python wrapper library, and a wrapper header file for integration with the Godot engine.

\subsection{Graphical Engine} Built upon the \Rulebook libraries, \Hammer incorporates an auto-configuring graphical engine developed using Godot. The Godot engine dynamically links to the compiled \Rulebook library (lib.so). At runtime, Godot scripts interpret and render the current game state, enabling modifications or extensions to \Rulebook libraries without requiring changes to graphical components.

The graphical engine can load and visually represent any valid game state and operates entirely client-side within a web browser.

Figure \ref{fig:two-side-by-side} demonstrates the graphical user interface provided by the 4Hammer graphical engine.

\subsection{Game Drivers} The final component of \Hammer comprises Game drivers, which load and run the game components independently of the graphical engine, allowing headless execution. Additionally, these drivers can connect over a network to control and manage execution within the graphical environment. The drivers are written in Python for ease of modification by the user.

\section{Experimental Results} 
\label{sec:eval}

The primary goal of our experimental validation is to demonstrate that the \Hammer environment is robust enough to serve as a reliable training environment. Consequently, we do not focus on creating a state-of-the-art machine learning model to master the environment. Instead, we use the \Rulebook compiler framework's built-in implementation of PPO~\cite{schulman2017proximalpolicyoptimizationalgorithms}, running on an Ubuntu machine equipped with an NVIDIA RTX 4070 GPU and an Intel Core i7-9700K CPU @ 3.60GHz to evaluate the headless mode. For validating the graphical engine driver, we employ Gemini 2 Flash.

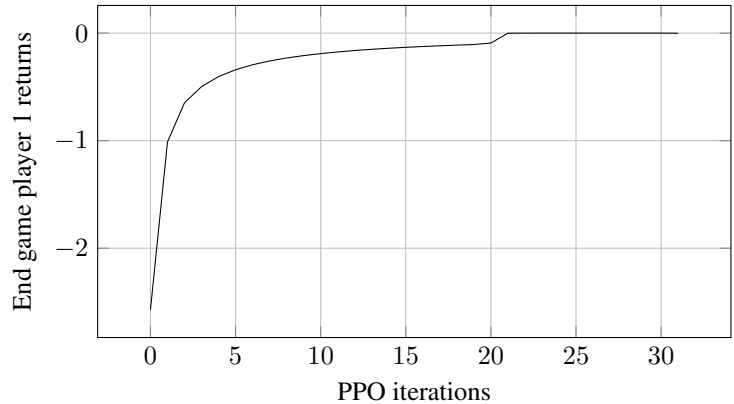
\begin{figure}
    \centering
    \begin{tikzpicture}
\begin{axis}[
    xlabel={PPO iterations},
    ylabel={End game player 1 returns},
    grid=major,
    width=10cm,
    height=6cm
]
\addplot[no markers] coordinates {
(0, -2.573298454284668)
(1, -1.0130000114440918)
(2, -0.6477707028388977)
(3, -0.4977973699569702)
(4, -0.40453460812568665)
(5, -0.3407035171985626)
(6, -0.29435601830482483)
(7, -0.2589763104915619)
(8, -0.23134668171405792)
(9, -0.20895829796791077)
(10, -0.19055649638175964)
(11, -0.1751033067703247)
(12, -0.16196846961975098)
(13, -0.1506889909505844)
(14, -0.1408587247133255)
(15, -0.13223247230052948)
(16, -0.1246170848608017)
(17, -0.11783107370138168)
(18, -0.11172141134738922)
(19, -0.1062362864613533)
(20, -0.09200000017881393)
(21, -0.0008999999845400453)
(22, -0.00039999998989515007)
(23, 0.0)
(24, 0.0)
(25, 0.0)
(26, 0.0)
(27, 0.0)
(28, 0.0)
(29, 0.0)
(30, 0.0)
(31, -0.001500000013038516)
};
\end{axis}
\end{tikzpicture}
    \caption{Average score obtained by the player 0 while training on single\_shooting\_maximize.rl}
    \label{fig:maximize}
\end{figure}

\paragraph{Headless Mode} We provide two example RL files describing simplified subsets of the full game. The first, \textbf{single\_shooting\_maximize.rl}, enables an agent to select one of two available units, with the selected unit subsequently attacking a third target unit. The agent's objective is to maximize inflicted damage. Figure \ref{fig:maximize} presents the average score achieved using the command \verb|rlc-learn|, with a learning rate set to 0.00001 and 1000 steps per environment.

\begin{figure} \centering 
\input{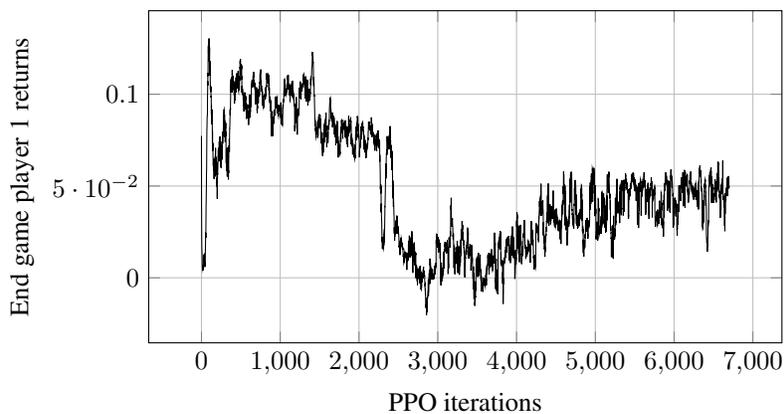}
\caption{Average score obtained by the agent during training on \texttt{single\_turn.rl}.} \label{fig:maximize2} \end{figure}

The second example, \textbf{single\_turn.rl}, simulates an entire game but limits each player to a single turn instead of the standard five. Unit selections are predetermined rather than chosen by the players. Additionally, the original scoring mechanism has been adjusted, as achieving significant score improvements within a single turn is not feasible. Instead, both players aim to maximize the difference in the number of remaining game pieces at the end of the match. Figure \ref{fig:maximize2} illustrates the average reward achieved during training, showing a significantly more irregular trend due to the direct impact of opponent actions on the player's score. These results were generated using rlc-learn with a learning rate of 0.00001, 5,000 steps per environment, and active league play.

\paragraph{Graphical Engine Mode}
To validate the graphical engine and Python driver mechanisms, we utilized Gemini 2 Flash\footnote{\href{gemini 2 flash}{https://cloud.google.com/vertex-ai/generative-ai/docs/models/gemini/2-0-flash}} to select actions. A python script keeps the internal state of the game and at each move provides the full textual serialized state to Gemini, along with all the valid moves in that state. Gemini selects a move among them, then the python script forwards the move to the graphical engine, displaying it on screen.

While Gemini actions were fairly poor, it was fairly competent at reading correctly the textual game state. As an example, in figure \ref{fig:log-output} we report the message that pushed Gemini to not reroll some dices when given the possibility.
\begin{figure}[h]
\centering
\begin{verbatim}
The game is in the Fight phase, specifically in the Charge phase where 
the von Ryan's Leapers (unit 6) are attempting to charge the Infernus 
Squad (unit 2). The current player is true (player 2 - tyranids). 
The von Ryan's Leapers are already eligible to charge. 
The question is whether to use any stratagems or abilities to 
influence the charge roll.

Given that the current roll is 3 + 4 = 7, which seems high enough to reach 
the target, and there's no visible way to improve this without spending 
command points,  the best action is to skip any further actions and let 
the charge proceed.
\end{verbatim}
\caption{Gemini 2 Flash inner monologue when considering if it should try rerolling a pair of dices. The decision required the model to understand which pieces belonged to the opponent and which did not, what was the decision to perform in the moment, the previous result that was being prompted to reroll, what was the cost of rerolling, and if that cost was worth it or not.}
\label{fig:log-output}
\end{figure}

\section{Conclusions} 
\label{sec:conc}

In this paper, we introduce \Hammer, a reinforcement learning environment specifically designed to test the performance of machine learning techniques on tasks in the field of board games with hour-long horizons. We presented the modular architecture of \Hammer, capable of accommodating diverse scenarios from traditional reinforcement learning methods to LLM-driven interaction over network protocols. We then experimentally validated our environment under conditions representative of its intended use cases.

Future work for \Hammer includes leveraging advanced reinforcement learning techniques to achieve superhuman-level gameplay performance.

\begin{ack}
This work is partially supported by the Italian Ministry of Enterprises and Made in Italy (MIMIT)
under the program “Accordi per l’innovazione nella filiera del settore automotive”, through the grant
"Piattaforma ed ecosistema cooperativo, C- ITS ETSI standard per la mobilità digitale
integrata", numero F/340043/01-04/X59, CUP B49J24001210005, finanziato a valere del Bando
MISE – ACCORDI PER L’INNOVAZIONE NEL SETTORE AUTOMOTIVE D.M. 31/12/2021 e DD
10/10/2022
\end{ack}

\bibliographystyle{plain}
\bibliography{bibliography}

%%%%%%%%%%%%%%%%%%%%%%%%%%%%%%%%%%%%%%%%%%%%%%%%%%%%%%%%%%%%

\appendix

%%%%%%%%%%%%%%%%%%%%%%%%%%%%%%%%%%%%%%%%%%%%%%%%%%%%%%%%%%%%

\end{document}